\documentclass[10pt,twocolumn,letterpaper]{article}

\usepackage[pagenumbers]{cvpr} 

\usepackage{graphicx}
\usepackage{amsmath}
\usepackage{amssymb}
\usepackage{booktabs}
\usepackage[accsupp]{axessibility}

\usepackage{algorithmic}
\usepackage{algorithm}
\usepackage{pifont}
\usepackage{graphicx}
\usepackage{bbm}
\usepackage{bbding}
\usepackage{multirow}
\def\Vec#1{{\boldsymbol{#1}}}

\usepackage[pagebackref,breaklinks,colorlinks]{hyperref}

\usepackage[capitalize]{cleveref}
\crefname{section}{Sec.}{Secs.}
\Crefname{section}{Section}{Sections}
\Crefname{table}{Table}{Tables}
\crefname{table}{Tab.}{Tabs.}

\def\cvprPaperID{1724} 
\def\confName{CVPR}
\def\confYear{2023}

\begin{document}

\title{Meta-causal Learning for Single Domain Generalization}

\author{
	Jin Chen\textsuperscript{\rm 1*},
	Zhi Gao\textsuperscript{\rm 1*},
	Xinxiao~Wu\textsuperscript{\rm 1,2\dag},
	Jiebo~Luo\textsuperscript{\rm 3}\\
	\textsuperscript{\rm 1}Beijing Key Laboratory of Intelligent Information Technology, \\ School of Computer Science \& Technology, Beijing Institute of Technology, China \\
	\textsuperscript{\rm 2}Guangdong Laboratory of Machine Perception and Intelligent Computing, \\
	Shenzhen MSU-BIT University, China \\
	\textsuperscript{\rm 3}Department of Computer Science, University of Rochester, Rochester NY 14627, USA\\
	{\tt\small \{chen\underline{\hspace{0.5em}}jin,gaozhi\underline{\hspace{0.5em}}2017,wuxinxiao\}@bit.edu.cn,jluo@cs.rochester.edu}
}
\maketitle

\renewcommand{\thefootnote}{\fnsymbol{footnote}} 
\footnotetext{$^*$ Jin Chen and Zhi Gao are co-first authors.}
\footnotetext{$^\dag$ Corresponding author: Xinxiao~Wu.}

\begin{abstract}
Single domain generalization aims to learn a model from a single training domain (source domain) and apply it to multiple unseen test domains (target domains). Existing methods focus on expanding the distribution of the training domain to cover the target domains, but without estimating the domain shift between the source and target domains. In this paper, we propose a new learning paradigm, namely \emph{simulate-analyze-reduce}, which first simulates the domain shift by building an auxiliary domain as the target domain, then learns to analyze the causes of domain shift, and finally learns to reduce the domain shift for model adaptation. 
Under this paradigm, we propose a meta-causal learning method to learn meta-knowledge, that is, how to infer the causes of domain shift between the auxiliary and source domains during training. We use the meta-knowledge to analyze the shift between the target and source domains during testing. 
Specifically, we perform multiple transformations on source data to generate the auxiliary domain, perform counterfactual inference to learn to discover the causal factors of the shift between the auxiliary and source domains, and incorporate the inferred causality into factor-aware domain alignments. 
Extensive experiments on several benchmarks of image classification show the effectiveness of our method.

\end{abstract}

\section{Introduction}
\label{sec:intro}

Single domain generalization~\cite{qiao2020learning} aims to generalize a model trained using one training domain (source domain) into multiple unseen test domains (target domains). Since only one source domain is given and the target domains are out-of-distribution and unavailable during training, single domain generalization is a challenging task and attracts increasing interests. Existing works have made considerable successes through expanding the distribution of the source domain by data augmentation~\cite{qiao2020learning,li2021progressive,wang2021learning} or learning adaptive data normalization~\cite{fan2021adversarially} typically.
However, such successes have been achieved without explicitly considering the domain shift between the source and target domains, which limits the generalization performance of model in real-world scenarios.  
\begin{figure}[!t]
	\centering
	\includegraphics[width=0.45\textwidth]{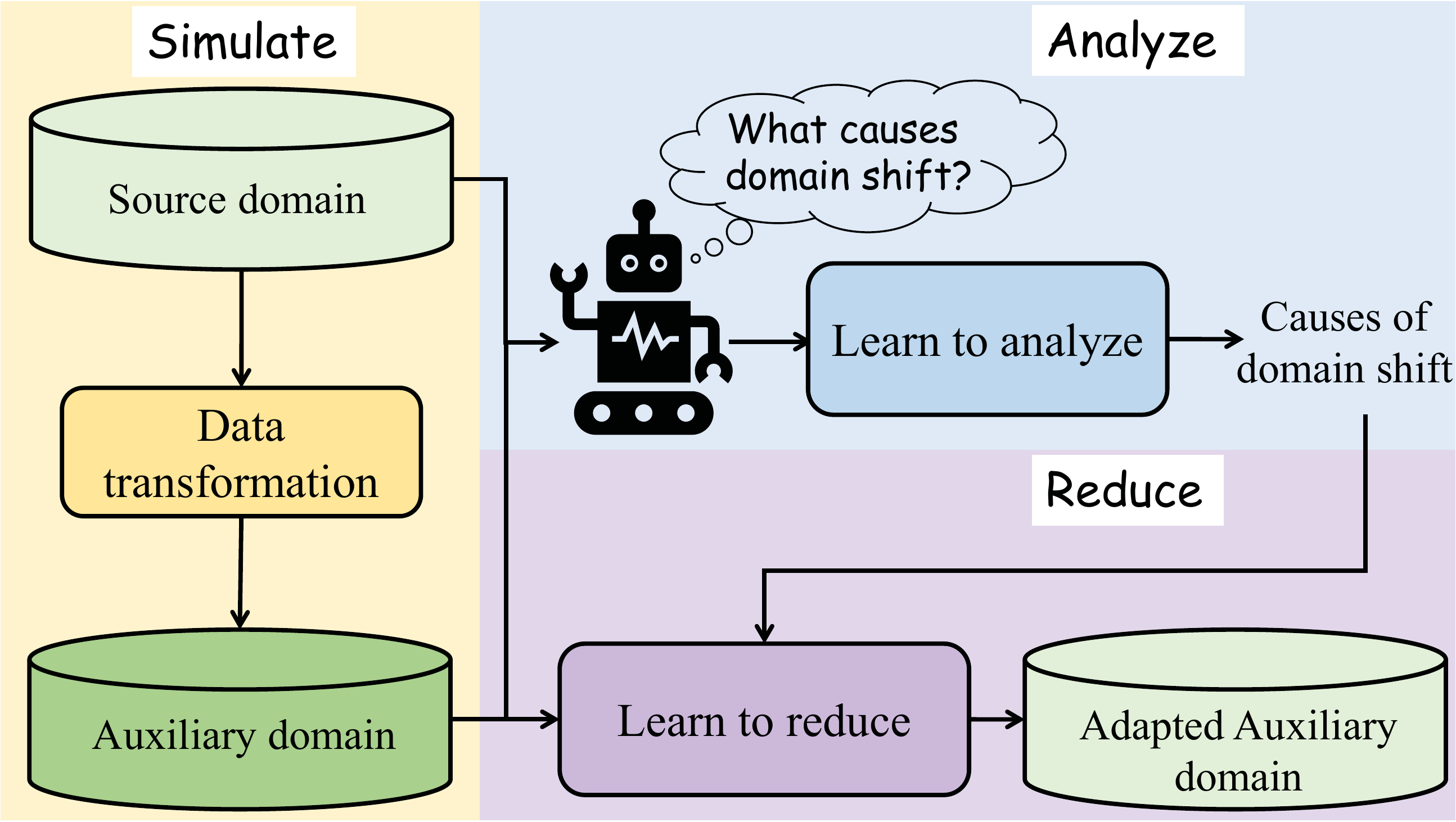}
	\caption{Illustration of the \emph{simulate-analyze-reduce} paradigm. In this paradigm, we first simulate the domain shift by constructing an auxiliary domain as the unseen target domain, then learn to analyze the domain shift, and finally learn to reduce the domain shift based on inferred causes.}
	\label{fig:paradigm}
\end{figure}

In this paper, we propose a new learning paradigm, namely~\emph{simulate-analyze-reduce}, to address single domain generalization by enabling the model to analyze the real domain shift between the source domain and unseen target domain. 
This new paradigm is shown in Figure~\ref{fig:paradigm}. We first build an auxiliary domain as the target domain to simulate the real domain shift between the source and target domains, since the target data is unavailable during training. We then learn to analyze the intrinsic causal factors of the domain shift to facilitate the subsequent model adaptation. Finally, we learn to reduce the domain shift with its inferred causes. 

Under this paradigm, we propose a meta-causal learning method to learn the meta-knowledge about how to infer the causes of the simulated domain shift between the auxiliary and source domains via causal inference in training, and then apply the meta-knowledge to analyze the real domain shift between the target and source domains during testing, through which the source and given target domains are adaptively aligned.
Specifically, we perform multiple transformations on source data to generate an auxiliary domain with great diversity.
Then we build a causal graph to represent the dependency among data, variant factors, semantic concepts and category labels, and conduct counterfactual inference over the causal graph to exploit the intrinsic causality of the simulated domain shift between the auxiliary and source domains. 
For each sample in the auxiliary domain, we construct  counterfactual scenes by intervening  variant factors to infer their causal effects  on the  category prediction, and these inferred causal effects of variant factors can be regarded as the causes of domain shift. To reduce the domain shift, we propose a factor-aware domain alignment by learning and integrating multiple feature mappings, where an effect-to-weight network is designed to convert the causal effects of variant factors into the weights of feature mappings.

During testing, the distribution discrepancy between the input target sample and the source domain is analyzed and reduced by applying the learnt meta-knowledge, \emph{i.e.,} inferring the causal effects of variant factors and incorporating them into the factor-aware domain alignment. 
In summary, the main contributions of this paper are as follows:
\begin{itemize}
	\item We propose a novel learning paradigm, \emph{simulate-analyze-reduce}, for single domain generalization. This paradigm empowers the model with the ability to estimate the domain shift between the source domain and unseen target domains, thus boosting the model adaptation across different domains.
	\item We propose a meta-causal learning method based on counterfactual inference to learn the meta-knowledge about analyzing the intrinsic causality of domain shift, thus facilitating the reduction of domain shift.
	\item Our method achieves the state-of-the-art results on several benchmarks of image classification, especially on the more challenging tasks with a large domain shift, clearly demonstrating the effectiveness of our method.
\end{itemize}
\section{Related Work}
\subsection{Domain Generalization}
Domain generalization focuses on generalizing a model learned from multiple source domains to the unseen target domain. 
The key difference between domain adaptation and domain generalization is that during training, domain adaptation leverages unlabelled target data while domain generalization has no access to the target domain. 
Existing domain generalization methods can be roughly divided into two categories: learning domain-invariant feature representation from multiple source domains~\cite{muandet2013domain,ghifary2015domain,motiian2017unified,WangHLX19,du2020learning} and generating diverse more samples via data augmentation~\cite{volpi2018generalizing,zhou2020learning,ShankarPCCJS18,carlucci2019domain}.

Recently, single domain generalization~\cite{qiao2020learning} has attracted growing attention, where only one source domain is available during training and the model is evaluated on multiple unseen target domains. A rich line of works employ data augmentation for generating out-of-domain samples to expand the distribution of the source domain~\cite{qiao2020learning,li2021progressive,wang2021learning}.  Qiao~\emph{et al.}~\cite{qiao2020learning} propose meta-learning based adversarial domain augmentation to generate samples. Li~\emph{et al.}~\cite{li2021progressive} propose a progressive domain expansion network to generate 
multiple domains progressively via simulating various photometric and geometric transforms by style transfer based generators. 
Wang~\emph{et al.}~\cite{wang2021learning} propose a style-complement module to generate diverse images with different styles. 
Fan~\emph{et al.}~\cite{fan2021adversarially} use data normalization for single domain generalization, where an adaptive normalization scheme is learned to be incorporated with adversarial domain augmentation to enhance the generalization of the model.

The aforementioned methods aim to generate the source data distribution as diverse as possible to cover unseen target domains. When the expanded source distribution does not approximate the target distribution, the performance may significantly degrade, since there still exists a domain gap between the source and target domains.
To address this problem, our method learns to analyze and reduce the domain shift by building an auxiliary domain during training. 

\begin{figure*}[h]
	\centering
	\includegraphics[width=0.92\textwidth]{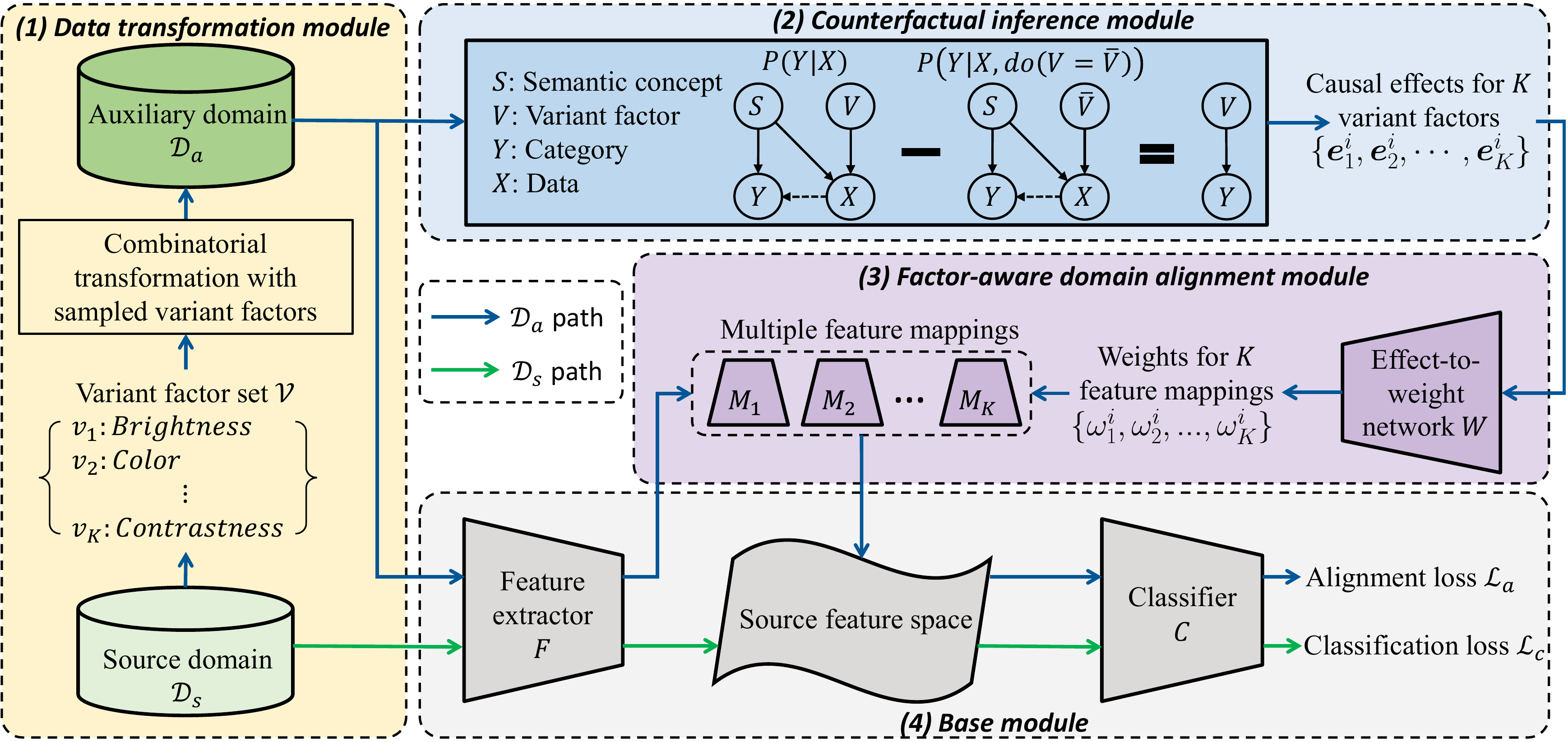}
	\caption{Overview of the proposed meta-causal learning method. (1) A data transformation module simulates the domain shift via generating an auxiliary domain $\mathcal{D}_a$. (2) A counterfactual inference module analyzes the domain shift by inferring the causal effects of $V$ on $Y$. ``\textbf{--}" denotes comparing the values of $Y$ before and after do-operation. The edge $X\to Y$ is a dashed arrow to represent the classification model $P(Y|X)$. (3) A factor-aware domain alignment module reduces the domain shift via multiple feature mappings according to weights learned by an effect-to-weight network. (4) A base module is used for feature extraction and classification.}
	\label{fig:framework}
\end{figure*}

\subsection{Causality for Domain Generalization}
Several recent methods exploit causality to learn domain-invariant semantic representation for domain generalization~\cite{liu2021learning,mahajan2021domain,lv2022causality}.
Considering the cross-domain invariance of the causality between semantic factors and predictions, Liu~\emph{et al.}~\cite{liu2021learning} propose a causal semantic generative model to remove the domain-specific correlation between semantic factors and variant factors, and thus make the prediction affected only by the semantic factors.
Assuming that images of the same object across domains should have the same representation, Mahajan~\emph{et al.}~\cite{mahajan2021domain} use the cross-domain invariance of the causality between objects and feature representations to capture the within-class variation for domain generalization. Lv~\emph{et al.}~\cite{lv2022causality} introduce causal inference to extract causal factors that are invariant across domains in order to learn invariant feature representation. 

From a different perspective, our method focuses on utilizing causal inference to discover the intrinsic causes of domain shift by constructing counterfactual scenes over the learnt causal graph, which facilitates the reduction of domain shift and in turn benefits cross-domain adaptation.

\section{Method}

\subsection{Problem Definition}
For single domain generalization, during training, we are given a labeled source domain $\mathcal{D}_s=\{(\Vec{x}_i^s,y_i^s)|_{i=1}^{N_s}\}$ drawn from the distribution $P_s$, where $\Vec{x}_i^s$ is the $i$-th source sample with its category label $y_i^s\in \mathcal{Y}$. 
During testing, the learnt model is applied to multiple unseen target domains $\mathcal{D}_t=\{\mathcal{D}_t^j\}_{j=1}^J$, where $\mathcal{D}_t^j$ is the $j$-th target domain drawn from the distribution $P_t^j$, and $P_t^j\neq P_s $. 

To bridge the domain gap between the source domain and multiple unseen target domains, we propose a new learning paradigm, called \emph{simulate-analyze-reduce}, which starts with simulating the domain shift between the source and target domains, then learns to analyze the domain shift, and finally learns to reduce the domain shift. Under this paradigm, we propose a meta-causal learning method that has four components: a data transformation module to generate auxiliary domains as target domains, a counterfactual inference module to discover the causes of the domain shift, a factor-aware domain alignment module to reduce the domain shift, and a base module for feature extraction and classification, as illustrated in Figure~\ref{fig:framework}.

\subsection{Data Transformation for Domain Shift Simulation}
\label{subsection:define}
To simulate the real domain shift between the source and target domains, we generate an auxiliary domain $\mathcal{D}_a$ as the unseen target domain by performing transformations on source data. The domain shift is actually the data distribution discrepancy between the source and auxiliary domains, and is usually caused by the variations of extrinsic attributes of data, independent of intrinsic semantics. Taking the image data for example, the domain shift is mainly caused by the variations of visual attributes, such as the variations of brightness, viewpoint, and color.  
Therefore, to make the simulated domain shift as realistic as possible, we formulate the extrinsic attributes as variant factors, and design data transformations according to the variant factors.

We define a set of variant factors, denoted as $\mathcal{V}=\{v_1, v_2,...,v_{K}\}$, 
where $v_k$ denotes the $k$-th variant factor. Each variant factor corresponds to a data transformation function that aims to generate new data by making changes on the corresponding extrinsic attribute, denoted as $G_{v_k}(\Vec{x};\theta_{v_k})$, where $\Vec{x}$ is an input sample, and $\theta_{v_k}\in [g_{min}^k,g_{max}^k]$ represents the degree parameter to control the magnitude of transformation with the scale range $[g_{min}^k,g_{max}^k]$. 

In real-world scenarios, the complex domain shift is often caused by  combinatorial multiple variant factors and accordingly we design a sampling strategy to enable the combinatorial data transformation.
Given a source sample $\Vec{x}_i^s$, we randomly sample $N_v^i$ variant factors from $\mathcal{V}$ to form a factor subset $\mathcal{V}_i=\{v_1^i,v_2^i, \cdots, v_{N_v^i}^i\}$.
Then, a corresponding auxiliary sample $\Vec{x}_i^a$ is generated by
\begin{equation}
	\label{equation:D_a}
	\begin{aligned}
		&\Vec{x}_i^a =  G_{v_{N_v^i}^i} \Big(\cdots G_{v_2^i} \big(G_{v_1^i} (\Vec{x}_i^s;\theta_{v_1^i});\theta_{v_2^i}\big)\cdots;\theta_{v_{N_v^i}^i}\Big),
	\end{aligned}
\end{equation}
where the transformation degree parameter $\theta_{v_k^i}$ is randomly selected from its range scale. In this way, the domain shift is generated as diverse as possible to approximate the real domain shift. 

\subsection{Counterfactual Inference for Domain Shift Analysis}
After simulating the domain shift, we introduce counterfactual inference to analyze the domain shift. During training, we learn the meta-knowledge about how to infer the causes of data discrepancy between one auxiliary sample and the source domain, and during testing, we apply the learnt meta-knowledge to unseen target samples.

We build a causal graph to model the causal dependency among the input sample (node $X$), the variant factors (node $V$), the semantic concepts (node $S$), and the output category (node $Y$), as shown in Figure~\ref{fig:framework}~(2).
The semantic concepts denote the intrinsic attributes of data that are related to the category, and the variant factors denote the extrinsic attributes of data that are domain specific, independent of the intrinsic semantics. For example, when the input sample is an image of ``zebra'', the semantic concepts are like ``four legs'' and ``black-white stripes'', and the variant factors include brightness, viewpoint, and so on. 
The edge $S\to Y$ represents that the category of the input data is determined by the semantic concepts. The edges $S\to X$ and $V\to X$ represent that the semantic concepts and the variant factors together determine what the input sample looks like. 
Since the semantic concepts represent the intrinsic semantics and are invariant across different domains, ideally the cross-domain classification is only determined by the semantic concepts, which is implemented by learning the edge $S\to Y$, \emph{i.e.,} estimating the conditional probability $P(Y|S)$. However, in reality, the semantic concepts are unobserved from the input samples, so the classification should be implemented by learning the edge $X\to Y$, \emph{i.e.,} estimating the conditional probability $P(Y|X)$. The edge $X\to Y$ is a dashed arrow to represent the classification model $P(Y|X)$. Since the input sample node $X$ is affected by the semantic concept node $S$ and the variant factor node $V$ together, the category node $Y$ is also affected by the variant factor node $V$ through edges $V\to X\to Y$. 
As the variant factors are domain-specific, their causal effects on the category lead to the domain shift. Hence, we infer the causal effects of the variant factors on the category prediction to discover the causes of the domain shift. 
That is to say, we infer the causal effects of the node $V$ on the node $Y$ as causes of domain shift, and then learn to reduce the domain shift based on its causes.

To infer the causal effects of the node $V$ on the node $Y$, we first learn the edge $X\to Y$ using the source data $\mathcal{D}_s$ by a classification loss:
\begin{equation}
	\label{equation:cls}
	\mathcal{L}_{c}=\mathbb{E}_{(\Vec{x}_i^s,y_i^s)\sim P_s}\left[- \sum_{u} \mathbb{I}_{u=y_i^s} \log C(F(\Vec{x}_i^s))_u\right],
\end{equation}
where $(\Vec{x}_i^s, y_i^s)\in \mathcal{D}_s$, $F$ is a feature extractor, $C$ is the classifier to output $|\mathcal{Y}|$ category probabilities, $C(\cdot)_u$ is the $u$-th element of $|\mathcal{Y}|$ category probabilities, and $\mathbb{I}_{u=y_i^s}$ is an indicator function, meaning that if $u=y_i^s$, the value of $\mathbb{I}_{u=y_i^s}$ is $1$ and $0$ otherwise.
Then, for each sample from the auxiliary domain, we construct a factual scene and multiple counterfactual scenes over the causal graph, so as to infer the causal effects of variant factors on the category prediction. 
Given an auxiliary sample $\Vec{x}_i^a\in \mathcal{D}_a$, its factual category is predicted by 
\begin{equation}
	\label{equation:facturalY}
	\Vec{y}_i^a = P(Y|X) =C(F(\Vec{x}_i^a)),
\end{equation}
where $\Vec{y}_i^a\in \mathbb{R}^{|\mathcal{Y}|}$ is a 
category probability vector, and represents the value of node $Y$. 
For each variant factor, we construct a counterfactual scene to infer its causal effect by doing intervention on the variant factor node $V$. Let $do(V=v_k)$ denote the intervention on the node $V$, which is implemented by changing  extrinsic attributes of data through the transformation $G_{v_k}$ with multiple degree parameters.  
Accordingly, the counterfactual category of  $\Vec{x}_i^a$ is predicted by 
\begin{equation}
	\begin{aligned}
	\label{equation:counterfactualY}
	\Vec{y}_{i,v_k}^a &= P\big(Y|X,do(V=v_k)\big)\\
	&=\frac{1}{|\mathcal{M}|}\sum\limits_{\theta_{v_{k}}\in \mathcal{M}}C\Big(F\big(G_{v_k}(\Vec{x}_i^a;\theta_{v_{k}})\big)\Big),
	\end{aligned}
\end{equation}
where $\mathcal{M}$ is a set of degree parameters, obtained by uniformly sampling  magnitudes of transformation $G_{v_k}$ from the scale range. 
By comparing the factual and  counterfactual category probabilities of $\Vec{x}_i^a$, the causal effect of the $k$-th variant factor is calculated by
\begin{equation}
	\begin{aligned}
	\label{equation:Ek}
	\Vec{e}_k^i = P(Y|X)-P\big(Y|X, do(V=v_k)\big) =  \Vec{y}_i^a - \Vec{y}_{i,v_k}^a.
	\end{aligned}
\end{equation}
The inferred causal effect represents how much contribution the corresponding variant factor makes to the domain shift, and the larger the causal effect is, the more seriously the domain shift is caused by the variant factor. 
\begin{figure*}[!t]
	\centering
	\includegraphics[width=0.95\textwidth]{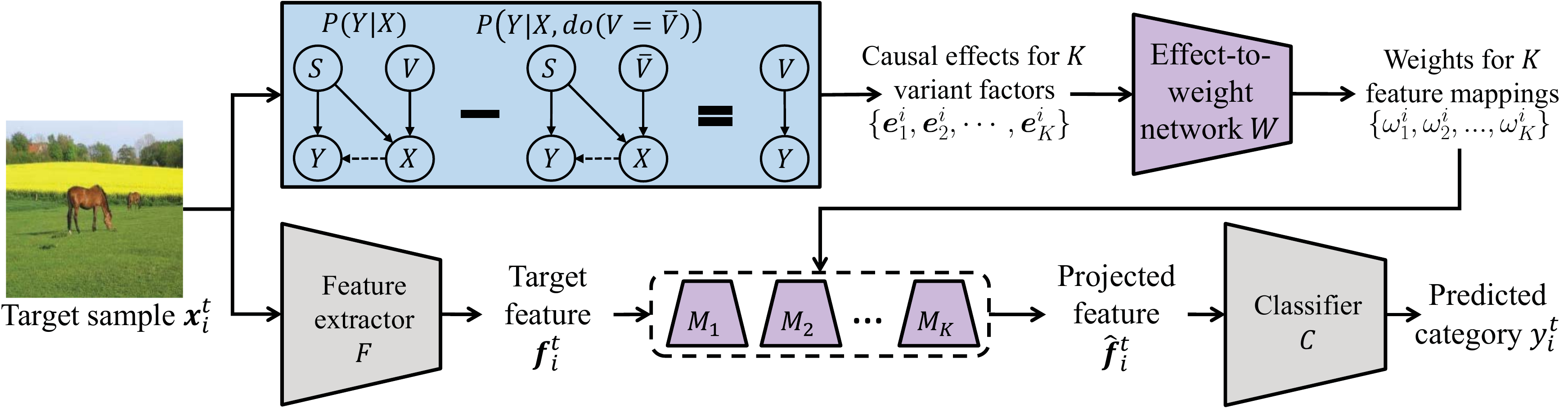}
	\caption{Inference process of meta-causal learning for a given target sample.}
	\label{fig:inference}
\end{figure*}

\subsection{Factor-aware Domain Alignment for Domain Shift Reduction}
After analyzing the causes of domain shift, we propose a factor-aware domain alignment to  reduce the domain shift by learning multiple feature mappings, with guidance of the inferred causal effects of variant factors. Each feature mapping addresses a specific domain shift caused by one variant factor. We construct $K$ feature mappings for $K$ variant factors, and the $k$-th feature mapping aims to address the domain shift caused by the $k$-th variant factor. In order to incorporate the causal effects of variant factors into the learning of mappings, we build an effect-to-weight network that converts the causal effect of each variant factor into the weight of the corresponding feature mapping.

For the auxiliary sample $\Vec{x}_i^a$ and its inferred causal effects of all variant factors \{$\Vec{e}_1^i,\Vec{e}_2^i,\cdots,\Vec{e}_K^i$\}, the weights of feature mappings are calculated by
\begin{equation}
	\label{equation:wk}
	\Vec{\omega}^i = {\rm{softmax}}\big(W(\Vec{e}_1^i),W(\Vec{e}_2^i),\cdots,W(\Vec{e}_K^i)\big),
\end{equation}
where $\Vec{\omega}^i \in \mathbb{R}^{K}$ and its $k$-th element $\omega^i_k$  denotes the weight of the $k$-th factor-aware feature mapping. $W(\cdot)$ denotes the effect-to-weight network. 

According to the weights of feature mappings, we integrate the $K$ feature mappings to project the auxiliary samples into the source feature space. The alignment of the source and auxiliary domains is implemented by minimizing the feature distance between the source and auxiliary samples in the source feature space. Given a source sample $\Vec{x}_i^s$ with the corresponding category label $y_i^s$, generated auxiliary sample $\Vec{x}_i^a $ and inferred mapping weights $\Vec{\omega}^i$, the alignment loss incorporated with inferred causal effects is defined as
\begin{equation}
	\begin{aligned}
		\mathcal{L}_{a}^{c}&=\frac{1}{N_s}\sum_i||F(\Vec{x}_i^s)-\sum_k \omega_k^i M_k\big(F(\Vec{x}_{i}^a)\big)||_2  \\
		&+\frac{1}{N_s}\sum_i \mathcal{H}\bigg(C\Big(\sum_k\omega_k^i M_k\big(F(\Vec{x}_i^a)\big),y_i^s\Big)\bigg).
	\end{aligned}
\end{equation}
where $F$ is the feature extractor, $M_k$ is the $k$-th feature mapping, $C$ is the classifier, $\omega_k^i$ is the $k$-th element of $\Vec{\omega}^i$,  $\sum_k \omega_k^i M_k\big(F(\Vec{x}_{i}^a)\big)$ represents the projected feature of $\Vec{x}_{i}^a$, and  $N_s$ is the number of source samples.  $||\cdot||_2$ is the L2-loss, and $\mathcal{H}(\cdot)$ is the cross-entropy loss.
The first term measures the feature distance between the source and auxiliary samples after projection, \emph{i.e.,} the data distribution discrepancy between the source and auxiliary domains. The second term encourages the auxiliary samples to belong to the same categories as the source samples.

Moreover, to enable each feature mapping to address the specific domain shift caused by the corresponding variant factor, we introduce another alignment loss $\mathcal{L}_{a}^{m}$:
\begin{equation}
	\begin{aligned}
		\mathcal{L}_{a}^{m}&=\frac{1}{N_s}\frac{1}{K}\sum_i\sum_k||F(\Vec{x}_i^s)- M_k\big(F(\Vec{x}_{i}^k)\big)||_2  \\
		&+\frac{1}{N_s}\frac{1}{K}\sum_i \sum_k \mathcal{H}\bigg(C\Big( M_k\big(F(\Vec{x}_i^k)\big),y_i^s\Big)\bigg),
	\end{aligned}
\end{equation}
where $\Vec{x}_i^k$ is an sample generated by conducting the data transformation $G_{v_k}$ on the source sample $\Vec{x}_i^s$, denoted by $\Vec{x}_i^k = G_{v_k}(\Vec{x}_i^s;\theta_{v_k})$. The transformation degree parameter $\theta_{v_k}$ is randomly selected from its range scale.

Then the overall loss function is defined as
\begin{equation}
	\label{equation:L}
	\begin{aligned}
		\mathcal{L} = \mathcal{L}_c+\mathcal{L}_a^{c}+\mathcal{L}_a^{m}.
	\end{aligned}
\end{equation}
The whole training process is summarized in Algorithm~\ref{algorithm:SAR}.

\begin{algorithm}
	\caption{Training process}
	\label{algorithm:SAR}
	\begin{algorithmic}[1]
		\renewcommand{\algorithmicrequire}{\textbf{Input:}}
		\REQUIRE The source domain $\mathcal{D}_s$, the variant factor set $\mathcal{V}$.
		\renewcommand{\algorithmicensure}{\textbf{Output:}}
		\ENSURE The feature extractor $F$, the classifier $C$, $K$ feature mappings $\{M_i\}_{i=1}^{K}$, the effect-to-weight network $W$.
		\STATE Initialize $F$, $C$, $\{M_i\}_{i=1}^{K}$, $W$;
		\WHILE {not converge}
		\STATE Sample an image $\Vec{x}_i^s$ from $\mathcal{D}_s$;
		\STATE Construct a factor subset $\mathcal{V}_i$ by sampling $N_v^i$ variant factors from $\mathcal{V}$;
		\STATE Generate an auxiliary sample $\Vec{x}_i^a$ by Eq.\eqref{equation:D_a};
		\FOR{$v_k$ in $\mathcal{V}$}
			\STATE Calculate the factual category $\Vec{y}_i^a$ of auxiliary sample $\Vec{x}_i^a$ by Eq.\eqref{equation:facturalY};
			\STATE Calculate the counterfactual category $\Vec{y}_{i,v_k}^a$ of auxiliary sample $\Vec{x}_i^a$ by Eq.\eqref{equation:counterfactualY};
			\STATE Infer the causal effect $\Vec{e}_k^i$ of variant factor $v_k$ via comparing $\Vec{y}_i^a$ and $\Vec{y}_{i,v_k}^a$ by Eq.\eqref{equation:Ek};
		\ENDFOR
		\STATE Calculate the weights of $K$ feature mappings $\{M_i\}_{i=1}^{K}$ by Eq.\eqref{equation:wk};
		\STATE Update $F$, $C$, $\{M_i\}_{i=1}^{K}$, $W$ by Eq.\eqref{equation:L}.
		\ENDWHILE
	\end{algorithmic}
\end{algorithm}

\subsection{Inference}
During testing, given a target sample $\Vec{x}_i^t$, firstly, the causal effects of variant factors are inferred by counterfactual inference in Eq.~\eqref{equation:Ek}, and the weights $\Vec{\omega^i}$ of feature mappings are calculated by the effect-to-weight network in Eq.~\eqref{equation:wk}. Then the target feature $\Vec{f}_i^t = F(\Vec{x}_i^t)$ is projected into the source feature space by integrating $K$ feature mappings according to their weights $\Vec{\omega^i}$ to obtain the projected feature $\hat{\Vec{f}}_i^t = \sum_k\omega_k^i M_k\big(\Vec{f}_i^t\big)$. Finally, the category of $\Vec{x}_i^t$ is predicted by $C(\hat{\Vec{f}}_i^t)$. The whole inference process is shown in Figure~\ref{fig:inference}.
\section{Experiments}
\subsection{Datasets}

\noindent\textbf{Digits.} The Digits dataset consists of five datasets: MNIST~\cite{lecun1998gradient}, MNIST-M~\cite{ganin2015unsupervised}, SVHN~\cite{netzer2011reading}, USPS~\cite{hull1994database}, and SYN~\cite{ganin2015unsupervised}, with $10$ categories. 
We use MNIST as the source domain, and the other four datasets as the target domains. The first $10,000$ images in the training set of MNIST are used for training.

\noindent\textbf{CIFAR10-C.} The CIFAR10-C dataset~\cite{HendrycksD19} is proposed to evaluate the robustness of classification model. The images are corrupted from the test set of the  CIFAR10 dataset~\cite{krizhevsky2009learning} by $19$ corruption types with five levels of severity. A higher level means the more serious corruption. There are $10$ categories. We use CIFAR10 as the source domain, and CIFAR10-C as the target domains where images of one severity level form one target domain. 

\noindent\textbf{PACS.} The PACS dataset~\cite{li2017deeper} is a benchmark for domain generalization, and consists of four domains: art painting, cartoon, photo, and sketch. There are $9,991$ images of seven categories. We use one domain as the source domain, and the rest three domains as the target domains. So there are four tasks with different domains as the source domains.

\subsection{Implementation Details}

\noindent\textbf{Auxiliary Domain.} 
We define $16$ variant factors to generate the images in the auxiliary domain, including $12$ photometric factors (\emph{Brightness, Contrast, Color, Sharpness, AutoContrast, Invert, Equalize, Solarize, SolarizeAdd, Posterize, NoiseSalt, NoiseGaussian}) and $4$ geometric factors (\emph{Shear-X, Shear-Y, Rotate, Flip}).
Since the \emph{Rotate} and \emph{Flip} variant factors will affect the semantic information of digit images, we use the other $14$ variant factors for the Digits dataset. For the CIFAR10-C and PACS datasets, all $16$ variant factors are used. To make the auxiliary domain as diverse as possible, for each source image at each iteration, we randomly sample several variant factors and use them to generate a new auxiliary image. 

\subsection{Results on Single Domain Generalization}

We compare our method with several state-of-the-art methods, including the baseline method (ERM~\cite{koltchinskii2011oracle}), the methods of learning domain-invariant features (CCSA~\cite{motiian2017unified}, d-SNE~\cite{xu2019d}, JiGen~\cite{carlucci2019domain}), and the methods of making data augmentation (GUD\cite{volpi2018generalizing}, M-ADA~\cite{qiao2020learning}, ME-ADA~\cite{Long2020Maximum}, PDEN~\cite{li2021progressive}, L2D~\cite{wang2021learning}, AA~\cite{cubuk2019autoaugment}, RA~\cite{cubuk2020randaugment}, RSDA~\cite{volpi2019addressing}, RSC~\cite{huang2020self}, ASR~\cite{fan2021adversarially}).
\begin{table}
	\caption{Single domain generalization results (\%) on Digits with ConvNet as backbone. The model is trained on MNIST, and evaluated on  SVHN, SYN, MNIST-M, and USPS.}
	\label{table:results_digits}
	\centering
	\resizebox{8.5cm}{!}{\begin{tabular}{l|cccc|c}
			\hline
			Method&SVHN&SYN&MNIST-M&USPS&Avg\\
			\hline
			ERM~\cite{koltchinskii2011oracle}&27.83&39.65&52.72&76.94&49.29\\
			\hline
			CCSA~\cite{motiian2017unified}&25.89&37.31&49.29&83.72&49.05\\
			d-SNE~\cite{xu2019d}&26.22&37.83&50.98&\textbf{93.16}&52.05\\
			JiGen~\cite{carlucci2019domain}&33.80&43.79&57.80&77.15&53.14\\
			\hline
			GUD~\cite{volpi2018generalizing}&35.51&45.32&60.41&77.26&54.62\\
			M-ADA~\cite{qiao2020learning}&42.55&48.95&67.94&78.53&59.49\\
			ME-ADA~\cite{Long2020Maximum}&42.56&50.39&63.27&81.04&59.32\\
			PDEN~\cite{li2021progressive}&62.21&69.39&82.20&85.26&74.77\\
			L2D~\cite{wang2021learning}&62.86&63.72&\textbf{87.30}&83.97&74.46\\
			AA~\cite{cubuk2019autoaugment}&45.23&64.52&60.53&80.62&62.72\\
			RA~\cite{cubuk2020randaugment}&54.77&59.60&74.05&77.33&66.44\\
			RSDA~\cite{volpi2019addressing}&47.40&62.00&81.50&83.10&68.50\\
			RSDA+ASR~\cite{fan2021adversarially}&52.80&64.50&80.80&82.40&70.10\\
			\hline
			Ours&\textbf{69.94}&\textbf{78.47}&78.34&88.54&\textbf{78.82}\\
			\hline
	\end{tabular}}
\end{table}
\begin{table}
	\caption{Single domain generalization results (\%) on CIFAR10-C with WRN as backbone. Each level is viewed as a target domain, a higher level denotes the more serious corruption and the domain discrepancy between the source and target domains is larger.}
	\label{table:results_cifar}
	\centering
	\resizebox{8.5cm}{!}{\begin{tabular}{l|ccccc|c}
		\hline
		Method&level1&level2&level3&level4&level5&Avg\\
		\hline
		ERM~\cite{koltchinskii2011oracle}&87.80&81.50&75.50&68.20&56.10&73.82\\
		\hline
		GUD~\cite{volpi2018generalizing}&88.30&83.50&77.60&70.60&58.30&75.66\\
		M-ADA~\cite{qiao2020learning}&90.50&86.80&82.50&76.40&65.60&80.36\\
		PDEN~\cite{li2021progressive}&90.62&88.91&87.03&83.71&77.47&85.55\\
		AA~\cite{cubuk2019autoaugment}&91.42&87.88&84.10&78.46&71.13&82.60\\
		RA~\cite{cubuk2020randaugment}&91.74&88.89&85.82&81.03&74.93&84.48\\
		\hline
		Ours&\textbf{92.38}&\textbf{91.22}&\textbf{89.88}&\textbf{87.73}&\textbf{84.52}&\textbf{89.15}\\	
		\hline
	\end{tabular}}
\end{table}

\begin{table}
	\centering
		\caption{Single domain generalization results (\%) on PACS with ResNet-18 as backbone. One domain (name in column) is used as the source domain and the other three domains are used as the target domains.} 
		\resizebox{8.5cm}{!}{\begin{tabular}{l|cccc|c}
			\hline
			Method&Artpaint&Cartoon&Sketch&Photo&Avg\\
			\hline
			ERM~\cite{koltchinskii2011oracle}&70.90&76.50&53.10&42.20&60.70\\
			\hline
			RSC~\cite{huang2020self}&73.40&75.90&56.20&41.60&61.80\\
			RSC+ASR~\cite{fan2021adversarially}&76.70&79.30&61.60&54.60&68.10\\
			\hline
			Ours&\textbf{77.13}&\textbf{80.14}&\textbf{62.55}&\textbf{59.60}&\textbf{69.86}\\
			\hline
		\end{tabular}
		\label{table:results_pacs}}
	\end{table}
\begin{table}
	\centering
	\caption{Leave-one-domain-out results (\%) on PACS with ResNet-18 as backbone. One domain (name in column) is used as the target domain and the other three domains are used as source domains.}
	\resizebox{8.5cm}{!}{
		\begin{tabular}{l|cccc|c}
			\hline
			Method&Artpaint&Cartoon&Photo&Sketch&Avg\\
			\hline
			MetaReg~\cite{balaji2018metareg}&83.70&77.20&95.50&70.30&81.70\\
			GUD~\cite{volpi2018generalizing}&78.32&77.65&95.61&74.21 &81.44\\
			Epi-FCR~\cite{li2019episodic}&82.10&77.00&93.90&73.00&81.50\\
			MASF~\cite{dou2019domain}&80.29&77.17&94.99&71.68&81.03\\
			JiGen~\cite{carlucci2019domain}&79.42&75.25&96.03&71.35 &80.51\\
			DMG~\cite{chattopadhyay2020learning}&76.90&80.38&93.55&75.21 &81.46\\
			DDAIG~\cite{zhou2020deep}&84.20 &78.10&95.30&74.70&83.10\\
            CSD~\cite{piratla2020efficient}&78.90&75.80&94.10&76.70&81.40\\
            L2A-OT~\cite{zhou2020learning}&83.30&78.20&96.20&73.60&82.80\\ 
            EISNet~\cite{wang2020learning}&81.89&76.44&95.93&74.33&82.15\\
            RSC~\cite{huang2020self}&83.43&80.31&95.99&80.85&85.15\\
			ME-ADA~\cite{Long2020Maximum}&78.61&78.65&95.57&75.59&82.10\\
			MMLD~\cite{matsuura2020domain}&81.28&77.16&96.09&72.29&81.83\\
			L2D~\cite{wang2021learning}&81.44 &79.56&95.51&80.58&84.27\\
			FACT~\cite{xu2021fourier}&85.37&78.38&95.15&79.15&84.51\\
			\hline
			MatchDG~\cite{mahajan2021domain}&81.32&80.70&\textbf{96.53}&79.72&84.57
			\\
			CIRL~\cite{lv2022causality}&\textbf{86.08}&80.59&95.93&82.67&86.32\\
			\hline
			Ours&85.30&\textbf{80.93}&\textbf{96.53}&\textbf{85.24}&\textbf{87.00}\\
			\hline
	\end{tabular}}
	\label{table:pacs_multi}
\end{table}
\begin{table}
	\centering
	\caption{Ablation study (\%) on PACS with ResNet-18 as backbone. One domain (name in column) is used as the source domain and the other three domains are used as target domains. ``T", ``A", ``C" denote Domain Transformation, Domain Alignment, and Counterfactual Inference, respectively.}
	\resizebox{8.5cm}{!}{
		\begin{tabular}{l|ccc|cccc|c}
			\hline
			Method&T&A&C&Artpaint&Cartoon&Sketch&Photo&Avg\\
			\hline
			Base&&&&71.26&67.64&43.97&36.99&54.97\\
			DT&\Checkmark&&&75.28&78.46&59.45&56.09&67.32\\
			DTA&\Checkmark&\Checkmark&&71.64&72.78&57.11&52.02&63.39\\
			\hline
			Ours&\Checkmark&\Checkmark&\Checkmark&\textbf{77.13}&\textbf{80.14}&\textbf{62.55}&\textbf{59.60}&\textbf{69.86}\\
			\hline
	\end{tabular}}
	\label{table:ablation_pacs}
\end{table}
	
Table~\ref{table:results_digits}, Table~\ref{table:results_cifar}, and Table~\ref{table:results_pacs} show the comparison results on Digits, CIFAR10-C, and PACS, respectively. 
From the results, there are several interesting observations as follows. First, our method generally outperforms the compared methods on all datasets, which clearly shows the effectiveness of the proposed simulate-analyze-reduce learning paradigm for single domain generalization.
Second, compared with the data augmentation methods (\emph{e.g.,} AA~\cite{cubuk2019autoaugment}, RA~\cite{cubuk2020randaugment} and RSDA~\cite{volpi2019addressing}) that are more related to our method, our method achieves much better results, and especially yields a $10.32\%$ gain over RSDA on Digits, strongly suggesting that it is beneficial to empower the model with the ability of analyzing the causes of domain shift by counterfactual inference. 
Third, on more difficult tasks with larger domain shift (\emph{e.g.,} SVHN on Digits, level5 on CIFAR10-C, and Photo on PACS), our method significantly improves the performance, further demonstrating the superiority of our method on handling more challenging situations. Forth, our method performs little worse on MNIST-M and USPS of Digits. The possible reason is that other compared methods use more well-designed data augmentation and network regularization, such as AdaIN based generators in PDEN~\cite{li2021progressive} and stochastic neighborhood embedding techniques in d-SNE~\cite{xu2019d}.

\subsection{Results on Multiple Domain Generalization}
We extend the proposed method to multi-source domain setting by regarding the multiple source domains as one source domain without using domain labels. We employ the leave-one-domain-out protocol following existing multi-source domain generalization~\cite{xu2021fourier,lv2022causality}.
We compare our method with most related methods that introduces causal inference into domain generalization (MatchDG~\cite{mahajan2021domain}, CIRL~\cite{lv2022causality}), and existing popular domain generalization methods (MetaReg~\cite{balaji2018metareg}, GUD~\cite{volpi2018generalizing}, Epi-FCR~\cite{li2019episodic},  MASF~\cite{dou2019domain}, JiGen~\cite{carlucci2019domain}, DMG~\cite{chattopadhyay2020learning}, DDAIG~\cite{zhou2020deep}, CSD~\cite{piratla2020efficient}, L2A-OT~\cite{zhou2020learning}, EISNet~\cite{wang2020learning}, RSC~\cite{huang2020self}, ME-ADA~\cite{Long2020Maximum}, MMLD~\cite{matsuura2020domain}, L2D~\cite{wang2021learning}, FACT~\cite{xu2021fourier}).

Table~\ref{table:pacs_multi} shows the leave-one-domain-out results on the PACS dataset with ResNet-18 as backbone. 
From the results, we make several observations.
First, it is noteworthy that our method achieves the state-of-the-art overall performance (``Avg") although our method is not designed for multi-source domain generalization, which further demonstrates that the proposed simulate-analyze-reduce learning paradigm not only benefits single domain generalization but also boosts multi-source domain generalization. Second, compared with the methods of introducing causal inference to learn domain-invariant features (MatchDG~\cite{mahajan2021domain}, CIRL~\cite{lv2022causality}), our method achieves better results on the overall metric ``Avg", especially making $5.52\%$ and $2.57\%$ gains over MatchDG~\cite{mahajan2021domain} and CIRL~\cite{lv2022causality}, respectively, on the more challenging task (Sketch$\to$Others). 
Such improvements are attributed to the ability of analyzing and reducing the domain shift, further demonstrating the advantages of causal inference in analyzing the causes of the domain shift.

\begin{figure*}[!t]
	\centering
	\includegraphics[width=0.93\textwidth]{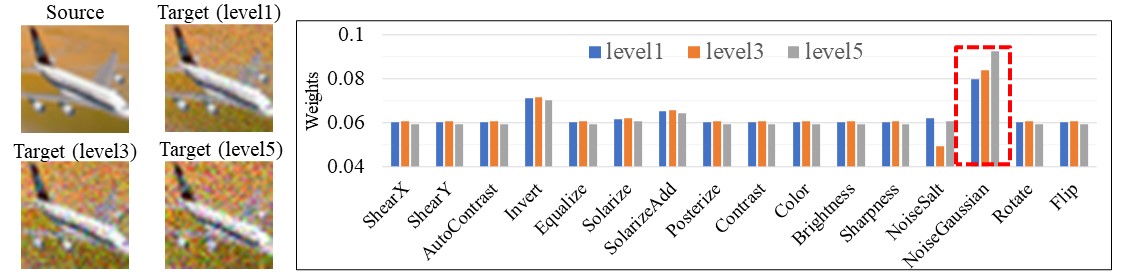}
	\caption{Examples of the inferred causal effects (represented as weights) of variant factors. The left part shows a source image from the CIFAR10 dataset and three target images from the CIFAR10-C dataset with Gaussian noise corruption. As the corruption severity increases from level 1 to level 5, the inferred weights of the \emph{NoiseGaussian} variant factor become larger accordingly.}
	\label{fig:weight}
\end{figure*}

\subsection{Ablation Studies}
To evaluate each component of our method, we conduct ablation experiments on the PACS dataset.
We design several degraded variants of our method for comparison: (1) ``Base", where only the base module is utilized and optimized by Eq.~\eqref{equation:cls} using the source domain; 
(2) ``DT'', where the data transformation module is added into the base module and the model is trained using both source and auxiliary domains;
(3) ``DTA'', where the simulated domain shift between auxiliary domain and source domain is directly reduced without analyzing the causes of the domain shift via the proposed counterfactual module.
	
The results are shown in Table~\ref{table:ablation_pacs}. From the results, we make several observations. First, our method achieves better performance than ``DT'', which validates the superiority of our method on analyzing and reducing the domain shift, rather than directly enlarging the source data distribution. 
Second, ``DTA" performs worse than ``DT'', probably because brute-force domain alignment without analyzing causes of the domain shift leads to negative transfer and thus hurts the performance. Third, our method achieves the best performance thanks to the proposed simulate-analyze-reduce paradigm. 
\begin{table}
	\centering
	\caption{Factor analysis (\%) on PACS with ResNet-18 as backbone. One domain (name in column) is used as the source domain and the other three domains are used as target domains.}
	\resizebox{8.5cm}{!}{
		\begin{tabular}{l|cccc|c}
			\hline
			Method&Artpaint&Cartoon&Sketch&Photo&Avg\\
			\hline
			Ours\_GT&74.62&69.78&52.74&43.34&60.12\\
			Ours\_PT&74.47&73.48&50.35&51.67&62.49\\
			\hline
			Ours&\textbf{77.13}&\textbf{80.14}&\textbf{62.55}&\textbf{59.60}&\textbf{69.86}\\
			\hline
	\end{tabular}}
	\label{table:factor_pacs}
\end{table}		
\subsection{Factor Analysis}	
In order to further analyze the effect of variant factors, we design several  variants of our method by using different variant factors, including only using geometric factors (``Ours\_GT''), and only using photometric factors (``Ours\_PT''). Since the factor number of the two type factors are different, we randomly select $4$ photometric factors to keep the same number as geometric factors and repeat experiments $10$ times to avoid the effect of sampling. The results are shown in Table~\ref{table:factor_pacs}. From the results, it is noteworthy that ``Ours\_PT" performs better than ``Ours\_GT", especially with the $8.33\%$ gain when using Photo as the source domain. The reason may be that the domain shift between Photo and the other target domains is mainly caused by photometric factors. Moreover, our method outperforms both ``Ours\_PT'' and ``Ours\_GT'', showing that both photometric factors and geometric factors are required to simulate the domain shift as diverse as possible.

\subsection{Causality Visualization}
In Figure~\ref{fig:weight}, we visualize the inferred causal effects (represented by the weights) of variant factors for the unseen target images  on the  CIFAR10-C dataset during testing. The target images are actually corrupted from the source images of CIFAR10 by Gaussian noise of five levels.
It is interesting to observe that the inferred weights of the \emph{NoiseGaussian} variant factor are larger than that of the other variant factors, indicating that the counterfactual inference succeeds in discovering the real cause of the domain shift.
We can also observe that when the corruption severity of Gaussian noise increases from level 1 to level 5 (\emph{i.e.,} the domain shift is more serious), the inferred weights of the \emph{NoiseGaussian} variant factor become larger, which further demonstrates that our method measures the magnitude of the domain shift correctly.

\section{Conclusion}
We have presented a new paradigm, \emph{simulate-analyze-reduce}, for single domain generalization. Our paradigm empowers the model with the ability to analyze the domain shift, instead of directly expanding the distribution of the source domain to cover unseen target domains. Under this paradigm, we have presented a meta-causal learning method that can learn meta-knowledge about inferring the causes of domain shift during training, and apply such meta-knowledge to reduce the domain shift for boosting adaptation during testing. 
Extensive experiments on several benchmark datasets have validated the effectiveness of the new learning paradigm and the advantage of meta-causal learning on analyzing the domain shift for domain generalization. 

\noindent\textbf{Acknowledgments.}
This work was supported in part by the Natural Science Foundation of China (NSFC) under Grant No 62072041.

{\small
\bibliographystyle{ieee_fullname}
\bibliography{egbib}
}

\end{document}